\documentclass[runningheads]{llncs}

\usepackage{eccv}

\usepackage{eccvabbrv}

\usepackage{graphicx}
\usepackage{booktabs}
\usepackage{newfloat}
\usepackage{listings}
\usepackage{multirow} 
\usepackage{amssymb}
\usepackage{subfig}
\usepackage{subcaption}
\usepackage{mwe}
\usepackage{amsmath}
\usepackage{overpic}
\usepackage{diagbox}
\usepackage[multiple]{footmisc}
\usepackage[accsupp]{axessibility}  

\usepackage{hyperref}
\usepackage{ulem}
\usepackage{orcidlink}

\makeatletter
\newcommand{\sf@counterlist}{}
\makeatother
\usepackage{tocbasic}

\begin{document}

\title{InfiniVerse: Occupancy Guided Unbounded Scene Generation for Autonomous Driving} 

\titlerunning{InfiniVerse}
\makeatletter
\newcommand{\printfnsymbol}[1]{%
  \textsuperscript{\@fnsymbol{#1}}%
}
\makeatother
\author{
Xiaoyu Ye\inst{1,4}\orcidlink{0009-0003-1187-1880}%
\thanks{Equal contribution.}%
\fnmsep
\thanks{This work was done during an internship at Huawei.} \and
Leheng Li\inst{2,4}%
\printfnsymbol{1}\fnmsep\printfnsymbol{2} \and
Xinyu Ji\inst{3}\orcidlink{0009-0004-7694-6609}%
\printfnsymbol{1} \and
Yingjie Cai\inst{4} \and
Hongda He\inst{5}\orcidlink{0009-0009-8349-6200} \and
Xu Yan\inst{4}%
\thanks{Corresponding authors.} \and
Guanyi Zhao\inst{4} \and
Ying-Cong Chen\inst{2} \and
Bingbing Liu\inst{4} \and
Shuguang Cui\inst{1}\orcidlink{0000-0003-2608-775X} \and
Zhen Li\inst{1}%
\printfnsymbol{3}
}
\authorrunning{X. Ye et al.}
\institute{
The Chinese University of Hong Kong, Shenzhen \\
\email{xiaoyuye1@link.cuhk.edu.cn}\\
\email{\{shuguangcui,lizhen\}@cuhk.edu.cn}
\and
The Hong Kong University of Science and Technology (Guangzhou)
\and
Nanyang Technological University
\and
Huawei Technologies Co., Ltd.
\and
University of New South Wales
}

\maketitle

\begin{abstract}
Generating realistic, controllable, and temporally coherent urban environments is a critical yet unresolved challenge in the autonomous driving community. 
In this paper, we introduce \textbf{InfiniVerse}, a unified pipeline for long-range, 2D–3D-aligned, and controllable synthesis of dynamic urban scenes from a single frame. 
In practice, our approach first reconstructs a 3D occupancy representation from the input multi-view frame. This representation serves as a foundation for autoregressive scene extension along arbitrary trajectories. 
Subsequently, a video diffusion model translates the coarse occupancy grid into realistic, spatiotemporally consistent video sequences.
Moreover, we propose a hierarchical \textit{sketch-and-refine} paradigm, in which the generated videos are re-projected as image-conditioned feedback to enhance the 3D occupancy representation, establishing cross-modal alignment and mutual enhancement between the visual and spatial domains.
Extensive evaluations on the Waymo Open Dataset and nuScenes demonstrate that InfiniVerse achieves state-of-the-art performance, with a FID of \textbf{6.4} and FVD of \textbf{67.97}, significantly outperforming existing benchmarks in both duration and stability.

\end{abstract}

\section{Introduction}
\label{sec:intro}

The generation of realistic and controllable synthetic environments is a fundamental challenge in computer vision, with critical applications in autonomous driving, embodied AI, and digital twins~\cite{gao2024cat3dcreate3dmultiview,kim2023neuralfieldldmscenegenerationhierarchical,liu2024pyramiddiffusionfine3d,zyrianov2024lidardmgenerativelidarsimulation,zhang2024urbanscenediffusionsemantic,shi2024mvdreammultiviewdiffusion3d,xu2024grmlargegaussianreconstruction}. In the context of autonomous driving, the ability to synthesize long-duration, multi-view video sequences is essential for training and validating perception and planning systems. However, generating long-horizon driving scenes remains a significant challenge. Unlike short-range clips, long-term generation requires maintaining strict temporal consistency, physical plausibility, and geometric stability over extended trajectories, where even minor errors can quickly compound into significant visual distortions or structural collapses.
 
\begin{figure}[t]
    \captionsetup{type=figure}
    \includegraphics[width=1\textwidth]{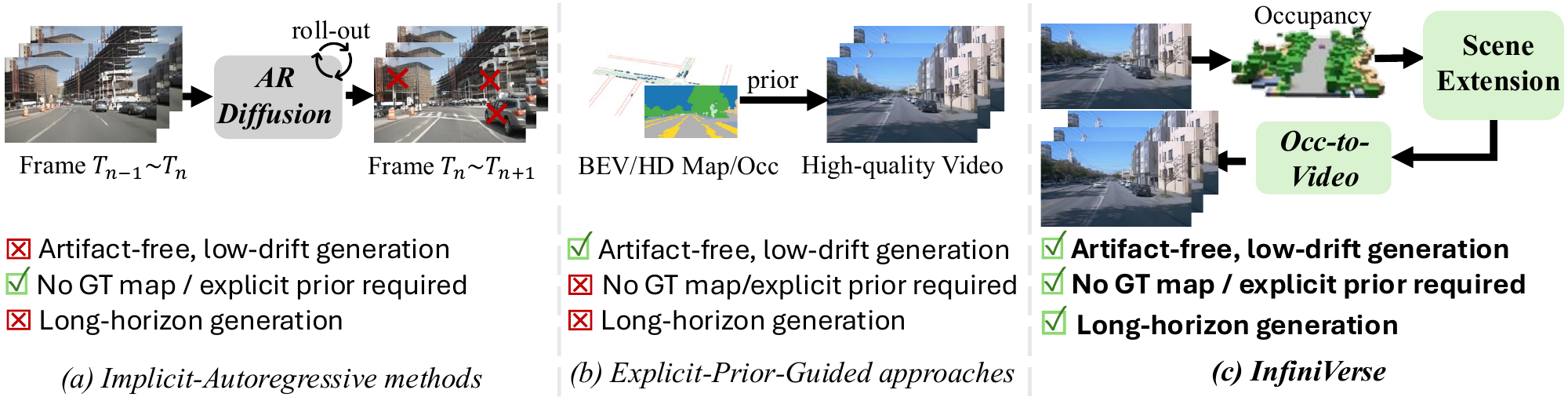}
    \caption{
\textbf{Paradigms for driving scene generation.}
Implicit video generation (a) produces visually realistic sequences but lacks explicit 3D structure, often causing geometric drift.
Geometry-driven pipelines (b) enforce spatial consistency but rely on strong priors such as BEV/HD maps or occupancy grids.
\textbf{InfiniVerse (c)} bridges these paradigms by constructing an extensible 3D occupancy world from a single multi-view observation and coupling it with video diffusion through a reciprocal refinement loop, enabling controllable and temporally consistent long-horizon driving video generation.
}
\label{fig:fig1}
\end{figure}

Existing generative approaches for driving scenes can be broadly categorized into two groups, yet both face limitations in long-horizon scenarios. 
The first category consists of \textbf{Implicit-Autoregressive methods}~\cite{wang2023drivingfuturemultiviewvisual, gao2024vistageneralizabledrivingworld, gao2023magicdrive}, which typically take a short video clip or a few frames as input and use video diffusion model to autoregressively predict future frames, as depicted in Fig.~\ref{fig:fig1}(a). While these models can produce high-fidelity appearance and plausible motion, they often rely on implicit latent features without an explicit intermediate geometric representation. Consequently, during long-term prediction, these methods are prone to accumulated errors and ``hallucinations'', where the generated content gradually deviates from the underlying 3D structure or loses spatial coherence. 
The second category focuses on \textbf{Explicit-Prior-Guided approaches}~\cite{lu2024wovogenworldvolumeawarediffusion, wang2023drivedreamerrealworlddrivenworldmodels, gao2025magicdrivev2highresolutionlongvideo, gao2025magicdrive3dcontrollable3dgeneration, wen2023panaceapanoramiccontrollablevideo, li2024uniscene, jiang2025diveefficientmultiviewdriving, ma2024unleashinggeneralizationendtoendautonomous} (\ie, Fig.~\ref{fig:fig1}(b)), requiring a specific BEV map/HD map as input. Although these provide better structural constraints, they often struggle to render long-horizon video due to the limited range of input prior. Furthermore, maintaining alignment between generated geometry and visual appearance remains a difficult open problem.

To address these limitations, we introduce \textbf{InfiniVerse} (see Fig. \ref{fig:fig1}(c)), a unified framework for long-range, controllable synthesis of dynamic urban environments from a single multi-view observation. Unlike prior works that rely on implicit transitions or rigid HD maps, our approach reconstructs an explicit 3D occupancy representation from the initial frame. This representation serves as a structural foundation for autoregressively extending the scene along arbitrary user-defined trajectories. To ensure high-fidelity results, we employ a video diffusion model to translate the extended occupancy grids into consistent multi-view video sequences.

The core of our method is a bidirectional ``sketch-and-refine'' loop that establishes a closed-loop feedback between the 3D occupancy and 2D video domains. Specifically, we treat the coarse occupancy as a ``sketch'' to guide video generation, and then re-project the generated high-fidelity frames back into the 3D space as image-conditioned feedback. By making 2D appearance and 3D geometry mutually constraining, our framework significantly reduces drift and suppresses hallucinations unsupported by the underlying scene structure. This reciprocal mechanism enables the creation of long-horizon, temporally consistent, and geometrically aligned driving scenarios.

In summary, our key contributions are:
\begin{itemize}
    \item We propose \textbf{InfiniVerse}, a unified pipeline that generates geometrically accurate and semantically coherent long-range 3D scenes from a single frame.
    \item We leverage a hierarchical \textbf{sketch-and-refine paradigm} that progressively refines the occupancy representation, enhancing scene consistency and controllability over long trajectories.
    \item Extensive experiments on the Waymo Open Dataset and nuScenes demonstrate that InfiniVerse achieves new state-of-the-art performance in both generation quality and duration.
\end{itemize}

\section{Related work}

\subsection{3D Urban Scene Generation}
Recent progress~\cite{ye2025hi3dgen} has been driven by learning-based methods that leverage powerful priors from large-scale 3D datasets, enabling diffusion model based 3D reconstructions~\cite{ren2024xcube, ren2024scube} even under severe ambiguity. Due to lack of high-quality 3D assets, several simulated environments played a vital role in autonomous driving, with platforms like CARLA~\cite{dosovitskiy2017carlaopenurbandriving} and Waymo’s Simulation Engine\cite{Sun_2020_CVPR_WOD} providing handcrafted 3D assets and rule-based control. However, these systems require significant manual effort and lack flexibility for arbitrary trajectory exploration. 

More recent methods attempt to automate urban scene generation by leveraging learned priors from real-world datasets. For instance, Infinicube~\cite{lu2024infinicube} has proposed techniques for large-scale scene synthesis, but they often depend on highly annotated HD maps and 3D layout bounding boxes, which limits scalability and data efficiency. X-Scene~\cite{yang2025xscenelargescaledrivingscene} explores large-scale driving scene generation guided by textual descriptions, enabling diverse scene composition and condition control through natural language prompts. However, its control mechanism remains coarse-grained and ambiguous, making it difficult to achieve precise spatial or semantic alignment between text instructions and generated content. Some methods~\cite{lee2024semcity, liu2024pyramiddiffusionfine3d, blockfusion} have exploited block-by-block generation for semantic scene generation in outdoor environments but lack controllability and fidelity.

\subsection{3D Occupancy Representation}
CityDreamer~\cite{xie2024citydreamercompositionalgenerativemodel} learns NeRF-style radiance fields or signed-distance functions to render novel views, but store geometry only implicitly; they cannot expose occupancy grids required for downstream planning or physical simulation. 

Occupancy grids have emerged as a powerful and compact representation for 3D scene understanding, especially in autonomous driving and robotics due to its explicit representation for efficient computation. They provide explicit volumetric reasoning and enable consistent spatial modeling across time and views. Works such as OccNet~\cite{sima2023_occnet}, VoxelNeXt~\cite{chen2023voxelnextfullysparsevoxelnet}, and Occ3D~\cite{tian2023occ3dlargescale3doccupancy} have shown promise in using learned occupancy fields for tasks like object detection and motion prediction. However, these approaches are typically limited to static or short-term settings and do not support long-range generation.

\subsection{Controllable Video Generation}
Video generation has recently seen significant progress with the introduction of scalable and general-purpose generative models. SVD (Stable Video Diffusion)~\cite{blattmann2023stablevideodiffusionscaling} introduces a diffusion-based framework that models video as a sequence of latent frames, enabling high-quality and temporally coherent video generation. Its scalable architecture and strong pretraining enable it to generalize well across diverse prompts and temporal lengths. Building on this, CogVideoX~\cite{yang2025cogvideoxtexttovideodiffusionmodels} proposes a diffusion transformer video generation framework that compresses videos across spatial and temporal dimensions in a 3D Variational Autoencoder (VAE). By using hierarchical spatiotemporal transformers and a factorized design, Large-scale diffusion models like VideoComposer~\cite{wang2023videocomposercompositionalvideosynthesis} and CogVideoX~\cite{yang2025cogvideoxtexttovideodiffusionmodels} condition on text, masks, or optical flow to produce plausible short clips, yet lack any notion of depth, occluded content, or kilometre-scale consistency but provide a very powerful base model.  
\begin{figure}[tb]
    \captionsetup{type=figure}
    \includegraphics[width=\textwidth]{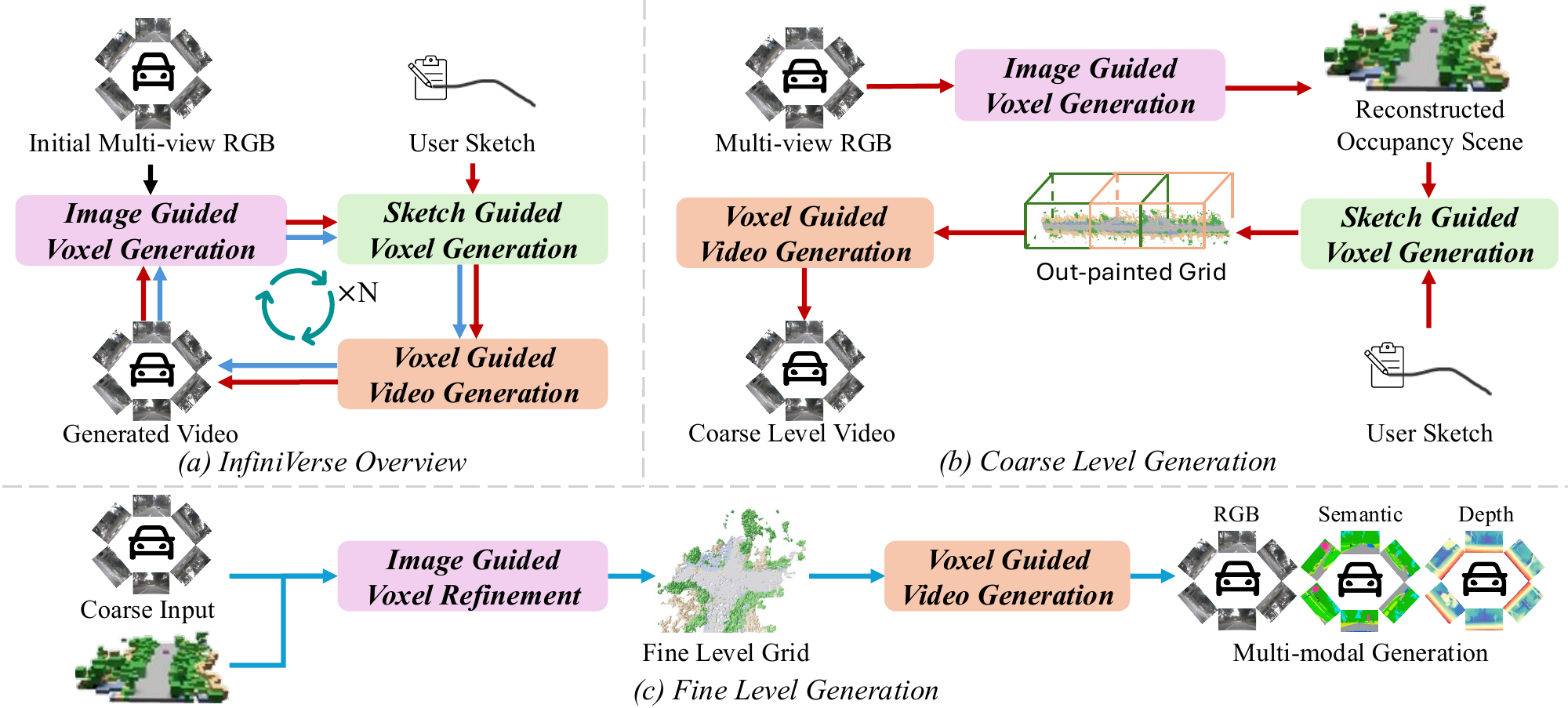}
    \captionof{figure}{\textbf{InfiniVerse overview.} We establish the system from a single multi-view frame by: 1) Constructing an image-guided voxel diffusion network to precisely reconstruct current initial occupancy scene and then encode the scene into triplane, performing sketch-guided occupancy generation to create a long-range, coarse level large-scale occupancy scene; and then 2) utilizing a fine-tuned video generator to translate the semantic world into coarse level driving video; 3) re-projecting the generated RGB video into image-conditioned voxel diffusion network to further refine each scene chunk, forming a reciprocal loop to provide 2D-3D aligned large scale scene generation. This scene can then be extended auto-regressively for long horizon exploration.}
    \label{pipeline}
\end{figure}

\section{Method}
\subsection{Overview}
Our framework adopts a two-stage design for large-scale 3D scene and long-term video generation. In the first stage, we reconstruct a detailed occupancy representation from a multi-view frame and encode it into a triplane-based generator, enabling fast and flexible synthesis of coarse 3D occupancy scene under arbitrary configurations. In the second stage, we fine-tune a video diffusion model to translate the synthesized occupancy scene into high-fidelity multi-view driving videos. The generated videos are further utilized to refine the occupancy representation, forming a fine-grained updating loop. Throughout the process, we construct reference images and multiple control conditions to ensure long-term temporal and cross-modal consistency between the occupancy representation and video generation.

\subsection{Image Guided Voxel Generation}
\label{occ recon}

We adopt XCube\cite{ren2024xcube} as the backbone for our initial scene reconstruction module. XCube is a large-scale 3D generative model that represents geometry using hierarchical sparse voxel grids and synthesizes scenes via a latent diffusion process. Specifically, it learns a distribution over a latent variable 
$\mathbf{X}$ encoded by a sparse structure Variational Autoencoder (VAE) defined on voxelized 3D geometry. The VAE encodes sparse voxel hierarchies into compact latent representations, and a diffusion model operates in this latent space to generate sparse voxel structures.
Both the VAE encoder–decoder and the diffusion model are implemented using sparse convolutional neural networks~\cite{graham2017submanifoldsparseconvolutionalnetworks}, enabling efficient modeling of high-resolution geometry up to $1024^3$.

 To condition XCube on input images, we use LSS~\cite{philion2020liftsplatshootencoding} unprojection pipeline to encode the image feature extracted from DINO-V2~\cite{oquab2024dinov2learningrobustvisual} into a dense 3D voxel grid $\mathbf{\Omega}$. The image features are processed with trainable 2D conv layers stored into feature channel $\mathbf{C}$
\begin{equation}
\mathbf{F}_{j d}^i=\theta_{j d}^i \cdot \mathbf{F}_j^i, \quad \mathbf{C}_v=\sum_{(i, j, d)} \mathbf{F}_{j d}^i \in \mathcal{R}^C .
\end{equation}
The $\theta_{j d}^i \in \mathbb{R}^D$ is $D$-dimensional Softmax-normalized vector pixel level depth distribution, the $d \in[1, D]$  denotes index of the depth pixel where $v$ denotes the index of a voxel. 
A key difference from indoor or object-centric setups is that outdoor driving scenes provide sparse and uneven camera coverage, we cannot naively broadcast per-pixel features along the entire ray, as done in prior work~\cite{liu2023one2345fastsingleimage, sun2021neuralreconrealtimecoherent3d, sitzmann2019deepvoxelslearningpersistent3d} since that the geometries is not precisely corresponding with camera frusta.
Once we constructed condition grid $\mathbf{C}$, we concatenate it with the latent $\mathbf{X}$ and feed into diffusion network as conditioning to produce high quality voxel grid for further extrapolator.
\label{sec:voxel recon}

\begin{figure}[tb]
    \captionsetup{type=figure}
    \includegraphics[width=\textwidth]{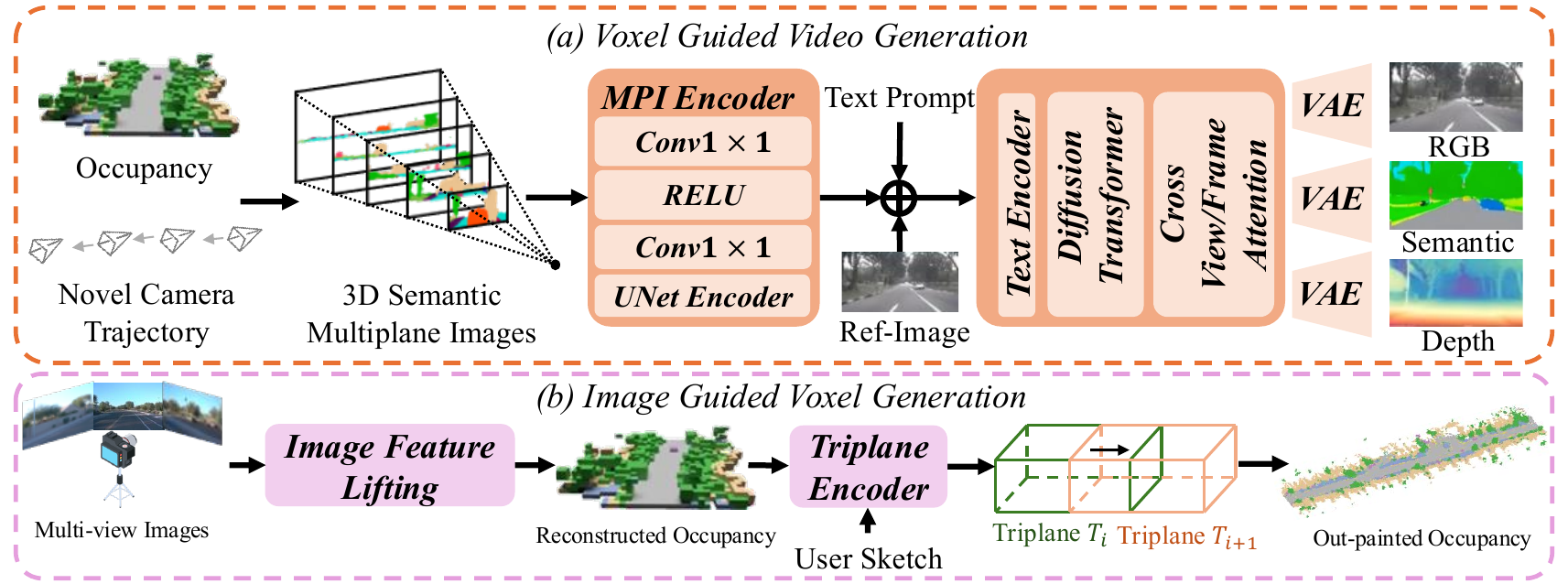}
    \captionof{figure}{Detail illustration on Voxel-to-Video diffusion and Video-to-Voxel diffusion. We Construct (a) by first encode the occupancy scene into Multiplane images to efficiently store both the semantic and geometric information, we designed a 1×1 convolutional encoder without downsampling to encode MPI features and transformed by a 1×1 convolution layer and ReLU activation. We also concatenate ref-image and Text conditions to further maintain the consistency and provide further control. With different VAE decoder we managed to output Multi-view RGB, semantics, depth video. As illustration (b), with a snapshot multi-view image, we first encode the image features in occupancy grid and with a Voxel diffusion, we reconstruct detailed occupancy scene and encode it in triplane with sketch condition $C_{sketch}$ for arbitrary sketch guided scene extrapolate. }
\label{Fig:Fig-3}
\end{figure}
\begin{figure}[tb]
    \captionsetup{type=figure}
    \includegraphics[width=\textwidth]{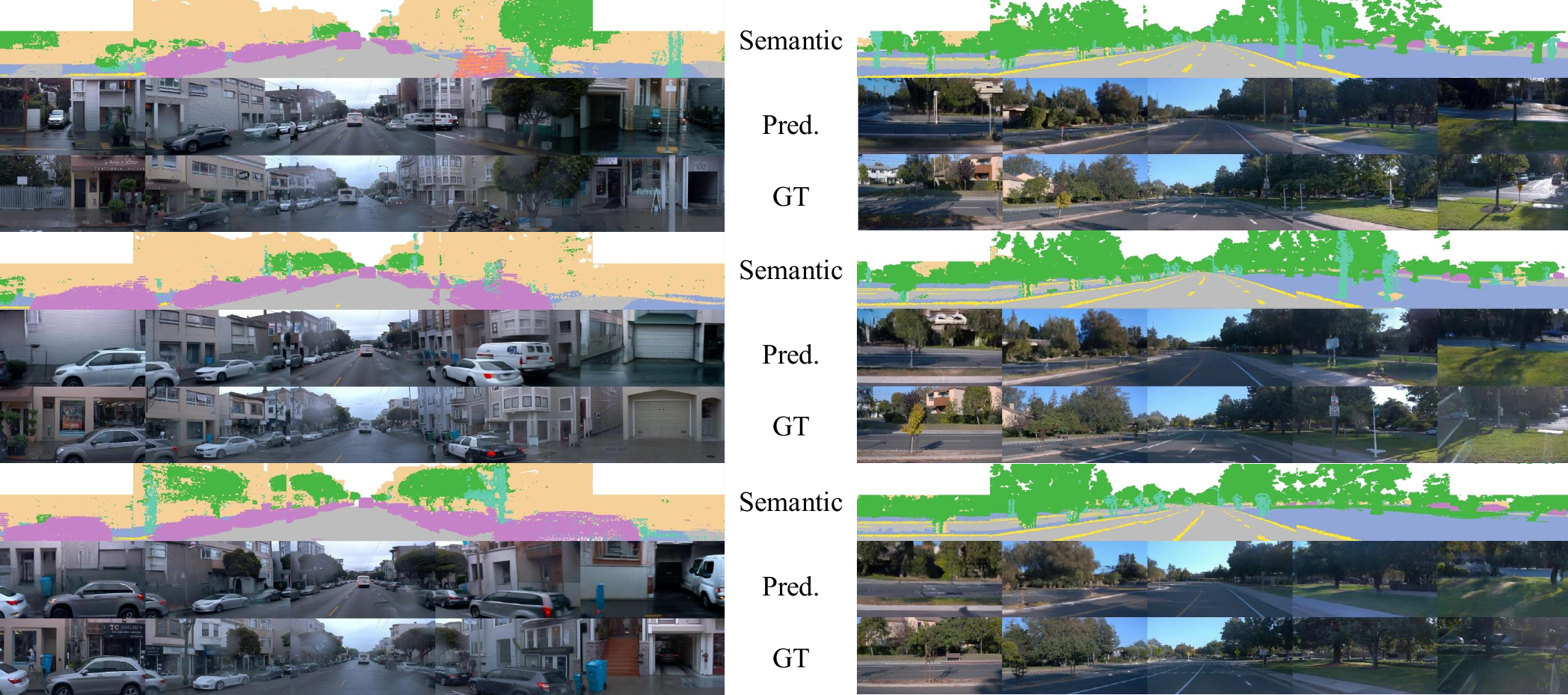}
    \captionof{figure}{Top to bottom: generated frames at T+5, T+15, and T+25 on Waymo Open Dataset. The Semantic row denotes Generated occupancy scenes encoded via the MPI encoder, while the ground-truth (GT) frames are shown in the bottom row. The comparison highlights the high fidelity and strong Temporal consistency of our generated sequences across the entire temporal span.}
\label{demo waymo}
\end{figure}

\subsection{Sketch Guided Voxel Generation}
Our framework represents the environment as an explicit occupancy grid $\mathcal{O}_t\in\{0,1\}^{H\times W\times D}$ that evolves over timesteps~$t$. Given the single-frame multi-view input, the reconstruction module (Sec. \ref{sec:voxel recon}) reconstructs $\mathcal{O}_0$ with fine-grained semantics and geometry. We regard this grid as the initial state of and learn a triplane based diffusion head $q(\mathcal{O}_{t+1}\mid\mathcal{O}_t,\mathbf{u}_t)$ that predicts the next occupancy volume conditioned on (i)~the current grid $\mathcal{O}_t$ and (ii)~a compact control vector $\mathbf{u}_t$ encoding the desired chunk Sketch trajectory.
To address the challenge of maintaining geometric and semantic coherence across extended trajectories, we propose a hierarchical sketch-and-refine paradigm that decomposes long-range occupancy generation into two stages: coarse-level sketching and fine-level image-conditioned refinement method.

To achieve controllable occupancy scene out-painting and fine-level image-conditioned occupancy updating, we encode the semantic occupancy grid to triplane latent representation~\cite{lee2024semcity}. The triplane latent projects the 3D occupancy onto three orthogonal rigid axes, decomposing the volumetric representation into three complementary 2D feature planes: $XY$, $XZ$, and $YZ$ planes. By querying features from all three planes and aggregating them through learned fusion mechanisms, we can regenerate detailed 3D semantic information while maintaining computational efficiency. In the subsequent parts of this section, all operations are performed within the triplane representation space.

\subsubsection*{Coarse-level Sketching Guided Generation.} Given an initial scene $\mathbf{O}_0$ and a trajectory sequence $\{T_j\}_{j=0}^N$, the sketch stage generates a coarse-level scene layout along the entire control trajectory. This stage operates at reduced spatial resolution to capture the global scene structure efficiently. The Sketch generation is conditioned on both the trajectory and the initial scene context for coherence:
\begin{equation}
\mathbf{X}_{sketch} = D_{\theta}(\mathbf{x}_0, \{T_j\}_{j=0}^N),
\end{equation}
Where $D_{\theta}$ denotes denoising loop, $\mathbf{X}_{sketch}$ represents the generated coarse-level scene spanning the full trajectory with a volume of $128^3$. 
Then we sample the occupancy alone the trajectory with a standard resolution triplane and also a sliding window denoising loop for a coarse level occupancy representation as shown here:
\begin{equation}
\mathbf{X}_{i} = D_\theta(\mathbf{V}_{i-1}; \mathbf{C}_{context}, \mathbf{C}_{sketch}),
\end{equation}
where $\mathbf{C}_{context}$ and $\mathbf{C}_{sketch}$ contains details from the previously refined patch $\mathbf{X}_{i-1}$, $\mathbf{V}_{i}$ indicates the triplane spatial rotation and translation vector for triplane to sample alone the trajectory. This step ensures smooth transitions and maintains local coherence across window boundaries.

This step provides a semantically consistent global layout that serves as the foundation for subsequent refinement.

\subsection{Bidirectional Closed-Loop Generation}
\subsubsection*{Controllable Video Generation.}
\begin{figure}[tb]
    \captionsetup{type=figure}
    \includegraphics[width=\textwidth]{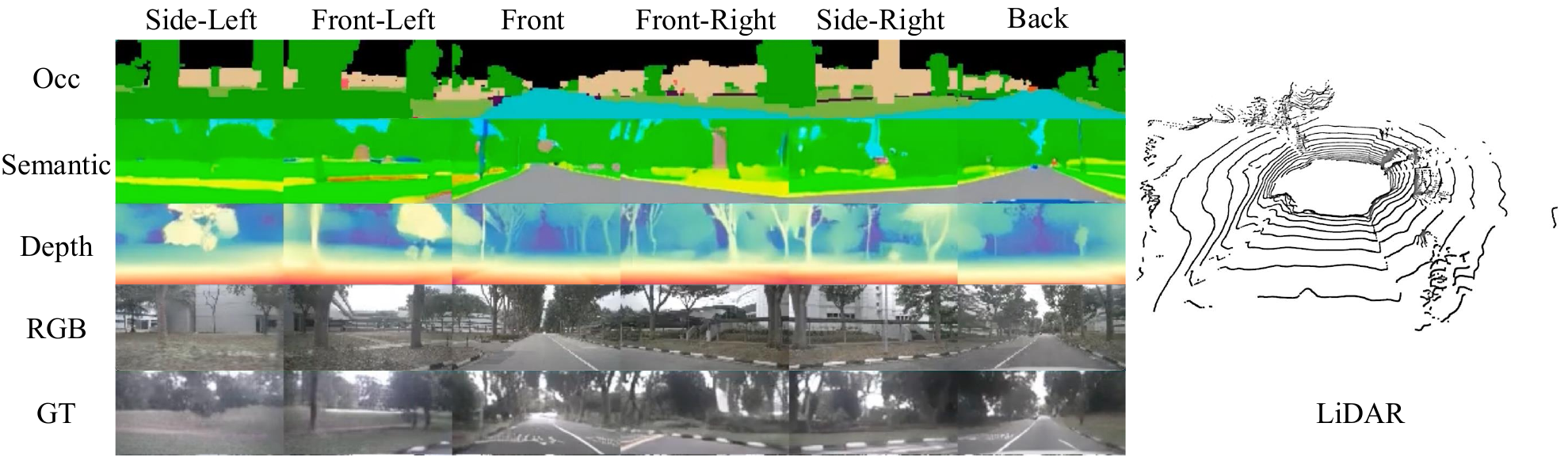}
    \captionof{figure}{Demonstration on aligned multiple multi-view sensor output compared with GT RGB Frame on nuScenes.}
\label{multi-modal nuscenes}
\end{figure}

We use CogvideoX1.5-5B as our base model and LoRA-fine-tuned on our processed dataset. It is non-trivial to design a 3D representation for conditioning and generating, but we aim to efficiently store both the semantic and geometric information of any occupancy input. We use multi-plane images (MPIs)~\cite{zhou2018stereomagnificationlearningview} as a representation to further improve video quality through acquiring 3D information of occluded objects.
\paragraph{ControlNet-style Injection.}
\label{controlNet}
As demonstrated in Fig. \ref{Fig:Fig-3}, we encode the $D$-plane stack acquired from camera frustum with a $1{\times}1$ Conv~$+$~ReLU encoder $\phi_M$ (no down-sampling) and inject it into the UNet decoder features $\{h^\ell\}$ at matching resolutions:
\begin{equation}
\tilde h^\ell \;\leftarrow\; h^\ell \;+\; \lambda_M\, \phi_M\!\big(\mathrm{MPI}^\ell\big),
\end{equation}
where $\lambda_M$ scales geometry control. Using $1{\times}1$ keeps exact spatial correspondence between multi-plane features and latent pixels, improving label alignment and compute over $3{\times}3$ alternatives.
\paragraph{Multi-view/frame Consistency.}
We use zero-initialized cross-view attention to exchange information with neighboring cameras and cross-frame attention to propagate temporal context:
\begin{align}
\mathrm{Attn}(Q,K,V)&=\mathrm{softmax}\!\Big(\tfrac{QK^\top}{\sqrt{d}}\Big)V,\\
h_{\text{out}}&=h_{\text{in}}+\!\!\sum_{i\in\{l,r\}}\!\!\mathrm{Attn}(Q_{\text{in}},K_i,V_i), \label{eq:xview}\\
h_{\text{out}}&=h_{\text{in}}+\!\!\sum_{i\in\{f,h\}}\!\!\mathrm{Attn}(Q_{\text{in}},K_i,V_i), \label{eq:xframe}
\end{align}
where $\{l,r\}$ denote left/right adjacent views and $\{f,h\}$ future/history frames. Eqs.~\eqref{eq:xview}--\eqref{eq:xframe} maintain instance coherence across overlapping FoVs and time. 
We also design a Ref-image initial conditioned injection mechanism for long-term video stability and maintain the temporal consistency. For every video generation chunk, we took last generated video frame or the initial frame if its the first chunk ${\mathrm{F}_t}$, and encoded it through vanilla VAE proposed from CogvideoX. Then we concat it through similar conditioned injection introduced in Sec \ref{controlNet} to formulated a text-pfirompted and ref-image enabled spatial temporal consistent video generation head. As illustrated in Fig. \ref{Fig:Fig-3} we route different vae decoders for various output.
Through this manner, we support 2D-3D aligned video generation with precisely occupancy-centric conditioning and outputs across semantics, depth and RGB demonstrated in Fig. \ref{multi-modal nuscenes} in nuScenes. We also present down-sampled lidar from our generated occupancy grid as lidar sensor output.

\subsubsection*{Fine-level Refinement.} The refinement stage operates on local scene patches using a sliding-window strategy to progressively enhance the coarse, sketch-based scene with fine-grained details contributed by the powerful video diffusion head. For each window position $i\text{-th}$, we encode the corresponding camera frustum into a multi-plane image (MPI), which serves as the 3D condition for a fine-tuned video diffusion model. This process generates spatially and temporally consistent multi-view videos with high visual fidelity.
With these high-quality videos, the large-scale driving scene generation task is now an image-conditioned occupancy reconstruction problem. We thus reuse the reconstruction model introduced in Sec. \ref{occ recon} to obtain detailed, high-resolution occupancy representations.
This improved occupancy further enables more accurate video generation, forming a closed-loop refinement process that progressively enhances both the 3D occupancy and the visual quality of the driving videos, achieving strong multi-modal alignment.

This two-stage approach enables the generation of arbitrarily long scenes while preserving both global semantic consistency through the sketch stage and local geometric fidelity through the refinement stage. The overlapping regions between adjacent windows provide continuity constraints that prevent artifacts and maintain seamless scene transitions throughout the extended trajectory.
\paragraph{Text-driven Weather and Style Control.}
High-level appearance is provided by a text encoder $\tau_\theta(\cdot)$ that embeds prompts $y$ into conditioning tokens $g=\tau_\theta(y)$. We form $y$ by concatenating weather/style clauses (e.g., ``snowy'', ``sandstorm'') with scene text; classifier-free guidance is used at sampling time. We inject $g$ via the standard cross-attention layers of the latent diffusion UNet.

\paragraph{Unified Denoising Objective.}
We train the video diffusion model using a standard latent denoising objective conditioned on all available control signals:
\begin{equation}
\mathcal{L}_{\text{base}} \;=\; 
\mathbb{E}_{\epsilon,\,t}\!\left[
\big\|\epsilon - \epsilon_\theta\!\left(z_t,\, t,\, \phi_M(\mathrm{MPI})\right)\big\|_2^2 \odot w
\right],
\end{equation}
where $\epsilon_\theta$ denotes the noise predicted by the denoising network at diffusion step $t$, $z_t$ is the noisy latent, and $\phi_M(\mathrm{MPI})$ encodes the multi-plane image (MPI) as the 3D control signal. The weight term $w$ optionally applies importance weighting, such as progressive or depth-aware weighting, to stabilize training and emphasize geometrically meaningful regions.

To further enhance the generation quality of dynamic scene components, we introduce an additional region-weighted reconstruction loss that assigns higher importance to road surfaces and road actors (vehicles, pedestrians, cyclists, et al.):
\begin{equation}
\mathcal{L}_{\text{region}} \;=\;
\mathbb{E}_{\epsilon,\,t}\!\left[
\big\| \big(\epsilon - \epsilon_\theta\!\left(z_t,\, t,\, \phi_M(\mathrm{MPI})\right)\big)
\odot m_{\text{road}}\big\|_2^2
\right],
\end{equation}
where $m_{\text{road}}$ is a binary or soft mask highlighting road and actor regions derived from semantic category.

The final training objective combines both terms:
\begin{equation}
\mathcal{L}_{\text{video}} \;=\;
\mathcal{L}_{\text{base}} \;+\; 
\lambda_{\text{region}}\,\mathcal{L}_{\text{region}},
\end{equation}
where $\lambda_{\text{region}}$ controls the relative importance of the region-weighted constraint.

This composite loss encourages the diffusion model to focus on both global temporal coherence and fine-grained generation of critical driving scene elements, yielding visually consistent and semantically accurate multi-view videos and also improves the fidelity of the assigned category in the bidirectional cross-Modal closed-loop generation.

\paragraph{Ultra-long video rollout.}
Let $\{x_{1:F}\}$ be the target frame sequence, and let $\mathcal{C}=\{\mathrm{MPI},g\}$ be the fixed controls for the rollout. We generate with a sliding-window sampler of length $W$ using cross-frame attention:
\begin{align}
p_\theta(x_{t}\mid x_{t-W:t-1},\mathcal{C}) \;\propto\; \prod_{k=K}^{1} p_\theta\!\big(z_{k-1}\mid z_k,\, \mathcal{C},\, x_{t-W:t-1}\big),
\end{align}
periodically re-anchoring features with $\phi_M(\mathrm{MPI})$ (every $R$ frames) to curb drift and preserve long-horizon geometric fidelity. The cross-frame operator in equation \eqref{eq:xframe} enables temporal propagation beyond training clip lengths; combined with prompt controls, it sustains consistent appearance over arbitrarily long sequences.

\section{Experiment}
\label{sec:exp}

\subsection{Dataset and Experimental Setup}

\paragraph{Data Curation}
We conduct comprehensive experiments on the nuScenes\cite{caesar2020nuscenesmultimodaldatasetautonomous} and Waymo Open Dataset~\cite{Sun_2020_CVPR_WOD}.
To support training and evaluation, we further extend the data processing pipeline provided by SCube~\cite{ren2024scube} to construct a more accurate 3D occupancy dataset. To enhance the geometric fidelity of our representations, we integrate dense geometry reconstructed by the multi-view stereo pipeline of COLMAP~\cite{schoenberger2016sfm}, which significantly enriches the voxel representation with fine-grained structural details and effectively addresses the height limitations inherent in LiDAR sensors. We further add dynamic road actors extracted from ground-truth LiDAR frames~\cite{tian2023occ3dlargescale3doccupancy}, including vehicles, pedestrians, and cyclists et al. A key focus of this data curation process is maintaining precise alignment across modalities—ensuring that RGB imagery, LiDAR point clouds, and occupancy representations are geometrically and temporally consistent within a unified world coordinate frame. This alignment is crucial for learning reliable 2D–3D correspondences, enabling InfiniVerse to jointly reason about appearance, geometry, and dynamics. Each curated scene therefore contains both static environmental geometry and temporally coherent dynamic elements, providing a strong foundation for long-horizon, controllable, and multi-modal grounded scene generation.
To enable precise text control for video generation, we implement a sophisticated multi-stage captioning process. First, we divide each video clip into three segments corresponding to different front-view perspectives. For each segment, we apply a multi-stage captioning process across three front-views, beginning with InternVL2-40B-AWQ~\cite{chen2024far,chen2024internvl} to generate dense video captions. These captions are then refined using Meta-Llama-3-8B-Instruct~\cite{llama3modelcard} to improve linguistic quality and ensure better alignment with the requirements of controllable video generation. To guarantee semantic consistency between video and text, we further apply VideoCLIP-XL~\cite{wang2024videoclipxladvancinglongdescription} to filter out poorly aligned video-caption pairs.

Following standard practices in the field, we remove low-quality voxel grids and sequences with predominantly static ego trajectories, resulting in a curated dataset of 618 sequences for training and 130 sequences for evaluation in Waymo Open Dataset. We also use Occ3D-nuScenes\cite{tian2023occ3dlargescale3doccupancy} to train the occupancy generation branch and nuScenes\cite{caesar2020nuscenesmultimodaldatasetautonomous} to train the multi-view multi-modal video generation branch. This filtering and dataset selection ensures the quality and relevance of our experimental data.

\begin{table}[tb]
\footnotesize
\centering
\caption{Comparison of input/output between previous method and ours. Our method produces aligned long video starts by only one frame of input. With minimal input we achieve a new state-of-the-art result in FID$(\downarrow)$  in nuScenes validation set. "MV video" denotes Multi-view video generation.}
\scalebox{0.9}{\begin{tabular}{l|c|cc|c|c}
\hline
\multirow{2}{*}{Method} &
  \multirow{2}{*}{Input Type} &
  \multicolumn{2}{c|}{Output Type} &
  \multirow{2}{*}{\begin{tabular}[c]{@{}c@{}}2D-3D \\ Aligned\end{tabular}} &
  \multirow{2}{*}{FID$\downarrow$}  \\ \cline{3-4}
 &
   &
  MV Video &
  ~~~~~~3D Repre.~~~~~~ &
   &
   \\ \hline

BEVGen\cite{swerdlow2024streetviewimagegenerationbirdseye} &
  BEV Map &
  $\times$ &
  $\times$ &
  - &
  40.48 \\
BEVControl\cite{yang2023bevcontrolaccuratelycontrollingstreetview} &
  BEV Sketch &
  $\times$ &
  $\times$ &
  - &
  29.04 \\
GenAD\cite{yang2024genadgeneralizedpredictivemodel} &
  Image+BEV map &
  $\times$ &
  $\times$ &
  - &
  23.90\\

Drive-WM\cite{wang2023drivingfuturemultiviewvisual} &
  Image+Action+Box/Map &
  \checkmark &
  $\times$ &
  - &
  15.8\\

DriveDreamer-2\cite{zhao2024drivedreamer2llmenhancedworldmodels} &
  Box+FoV Map &
  \checkmark &
  HD Map+BEV &
  $\times$ &
  25.0\\
MagicDrive\cite{gao2023magicdrive} &
  Pose+Box+BEV &
  \checkmark &
  $\times$ &
  - &
  16.59\\
MagicDrive-V2\cite{gao2025magicdrivev2highresolutionlongvideo} &
  Traj.+Box+BEV &
  \checkmark &
  $\times$ &
  - &
  20.91\\
MagicDrive-3D\cite{gao2025magicdrive3dcontrollable3dgeneration} &
  Traj.+Box+BEV &
  \checkmark &
  3DGS &
  $\times$ &
  20.67\\
InfiniCity\cite{lin2023infinicityinfinitescalecitysynthesis}&
  Voxel &
  $\times$ &
  Voxel &
  $\times$ &
  77.0\\
Panacea\cite{wen2023panaceapanoramiccontrollablevideo}&
  Image+Box+FoV Map &
  \checkmark &
  $\times$ &
  - &
  16.87\\
Vista\cite{gao2024vistageneralizabledrivingworld}&
  Action &
   $\times$&
  $\times$ &
  - &
  13.97\\
Uniscene\cite{li2024uniscene}&
  BEV &
  \checkmark &
  Voxel+LiDAR &
  $\times$ &
  6.45\\
DiVE\cite{jiang2025diveefficientmultiviewdriving}&
  Traj.+Box+FoV&
  \checkmark &
  $\times$ &
  - &
  10.68\\
DelPhi\cite{ma2024unleashinggeneralizationendtoendautonomous}&
  BEV Map &
  \checkmark &
  $\times$ &
  - &
  15.08\\
   X-Scene\cite{yang2025xscenelargescaledrivingscene} &
  Text+Box+HD Map &
  \checkmark &
  Voxel &
  $\times$ &
  12.77\\
InfiniCube\cite{lu2024infinicube}&
  Box+HD Map &
  $\times$ &
  Voxel+3DGS &
  $\times$ &
  80.13\\
  DriveGan\cite{kim2021drivegancontrollablehighqualityneural} &
  Image+Action &
  $\times$ &
  $\times$ &
  - &
  73.4 \\
  WovoGen\cite{lu2024wovogenworldvolumeawarediffusion}&
  Image+HD Map+Voxel &
  \checkmark &
  HD Map+Voxel &
  $\times$ &
  27.6\\
  DriveDreamer\cite{wang2023drivedreamerrealworlddrivenworldmodels} &
  Image+Action+FoV Map &
  \checkmark &
  $\times$ &
  - &
  14.9\\
  \hline
\textbf{InfiniVerse (ours)} &
  \textbf{Image} &
  \textbf{\checkmark} &
  Voxel+LiDAR &
  \textbf{\checkmark} &
  \textbf{6.40}\\ \hline
\end{tabular}
}
\label{tab: main}
\end{table}

\begin{figure}[b]
    \captionsetup{type=figure}
    \includegraphics[width=\textwidth]{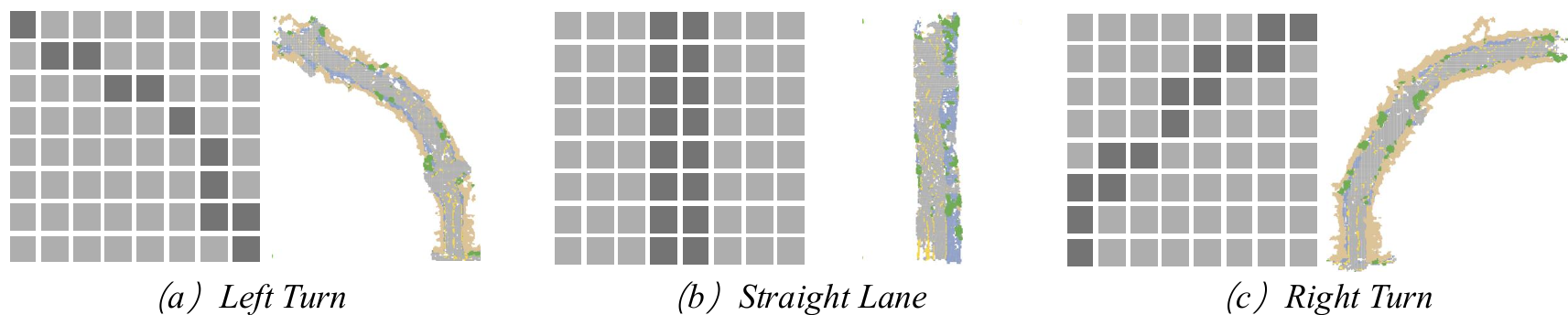}
    \captionof{figure}{Conditions and results for sketch guided voxel generation, we provide generated left turn, straight lane and right turn scenario based on corresponding sketch.}
\label{Fig:sketch}
\end{figure}

\subsection{Implementation Details}

Our implementation follows a carefully designed training protocol to ensure optimal performance across all components. The image-conditioned occupancy generation stage is trained for 100 GPU days including VAE compression, with a inference time in 2 minutes and 40GB VRAM per chunk. We use four 2D convolutional layers(channel dims: [768, 256, 256, 32, 32], kernel size: 3, stride: 1) to further process the DINO-V2 output to predict the image feature. The triplane based sketch guided diffusion is trained for 20 GPU days with inference time in 30s per triplane. For the triplane autoencoder, the input scene is encoded to triplane with a spatial resolution $(X_h, Y_h, Z_h) = (128, 128, 32)$, and the feature dimension $C_h$ is 16, the learning rate is initialized to 1e-4 and then decreases linearly. During the diffusion process, we use the default settings\cite{rombach2022highresolutionimagesynthesislatent} with 100 time steps ($T$). The video generation component is trained 64 GPU days, with a inference time in 2.5 minutes with 55GB VRAM per chunk. We adopt CogvideoX1.5-5B as our base model and LoRA-fine-tuned on our processed dataset. We use UniPC scheduler~\cite{zhao2023unipcunifiedpredictorcorrectorframework} with the classifier-free guidance (CFG)~\cite{ho2022classifierfreediffusionguidance} that is set as 7.0. During inference, we use 20 denoising steps for dataset generation. We employ a temporal compression ratio of 4 to balance computational efficiency with temporal fidelity. All experiments are conducted on NVIDIA A800 GPUs with 80GB VRAM with mixed precision training to optimize memory usage and training speed.
\begin{table}[]
\caption{Ablation of architectural components for video generation. We analyze the contribution of the 3D branch, the 2D--3D reciprocal loop, and the region-aware loss under the presence of a reference image. Without 3D branch the experiment is conducted under pure I2V mode. FID$(\downarrow)$ and FVD$(\downarrow)$ demonstrate that each component significantly improve generation quality.}
\centering
\begin{tabular}{{c|c|c|c|c|c}}
\hline
\multicolumn{1}{l|}{Ref-Image} &
  \multicolumn{1}{l|}{3D branch} &
  \multicolumn{1}{l|}{Reciprocal Loop} &
  \multicolumn{1}{l|}{Region Loss} &
  \multicolumn{1}{l|}{FID$\downarrow$} &
  \multicolumn{1}{l}{FVD$\downarrow$} \\ \hline
 $\times$&  \checkmark& $\times$ & $\times$ &       40.87        &      179.4          \\
 \checkmark& $\times$ & $\times$ & $\times$ &           47.89    &          320.6      \\
 \checkmark&  \checkmark&  $\times$& $\times$ &       19.12        &        142.12        \\
 \checkmark&  \checkmark&  \checkmark&  $\times$&         7.98      &       80.87         \\ \hline
 \textbf{\checkmark}&  \textbf{\checkmark}&  \textbf{\checkmark}&  \textbf{\checkmark}& \textbf{6.40} & \textbf{67.97} \\ \hline
\end{tabular}
\label{ablation framework}
\end{table}

\begin{table}[t]

\centering
\caption{FVD under different rollout lengths. Lower is better.}
\label{tab:fvd_vista}
\begin{tabular}{cccccc}
\toprule
\diagbox{Method}{Frames} & 12 & 24 & 36 & 48 & 96 \\
\midrule
Vista   & 78.4 & 86.2 & 94.7 & 105.9 & 141.3 \\
Ours    & \textbf{52.1} & \textbf{59.8} & \textbf{67.5} & \textbf{73.9} & \textbf{86.5} \\
\bottomrule
\end{tabular}
\end{table}

\begin{figure}[t]
    \centering
    \begin{subfigure}[t]{0.48\linewidth}
        \centering
        \includegraphics[width=\linewidth]{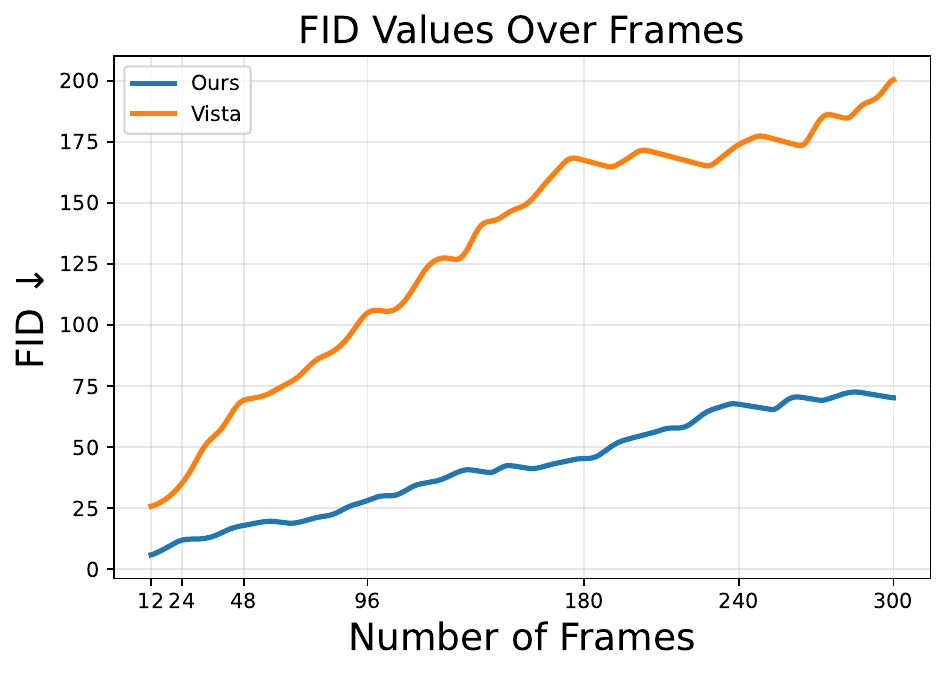}
        \caption{FID values over frames on Waymo.}
        \label{fig:fid_vs_frames}
    \end{subfigure}
    \hfill
    \begin{subfigure}[t]{0.48\linewidth}
        \centering
        \includegraphics[width=\linewidth]{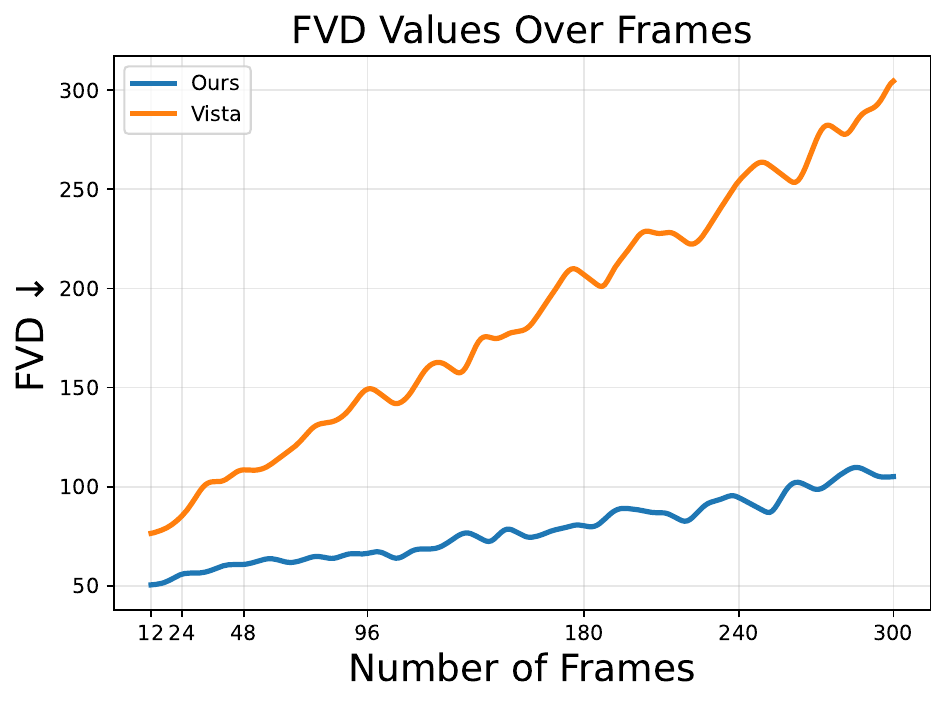}
        \caption{FVD values over frames on nuScenes.}
        \label{fig:fvd_vs_frames}
    \end{subfigure}
    \caption{Long-horizon generation quality under increasing rollout horizons. InfiniVerse shows a substantially slower degradation trend than Vista in both frame-level fidelity and temporal consistency.}
    \label{fig:fid_fvd_curves}
\end{figure}


\subsection{Metrics}
We use Fr\'echet Inception Distance (FID) ~\cite{heusel2018ganstrainedtimescaleupdate} and Fr\'echet Video Distance (FVD)\cite{DBLP:conf/iclr/UnterthinerSKMM19} to measure the perceptual quality of generated video, and use mIoU to measure the precision of occupancy prediction.

For Voxel grid reconstruction, we utilize IoU and mIoU, the final IoU of the fine-level voxel grid reached 34.31\%, and the mIoU that considers the accuracy of the voxel’s semantic prediction reached 20.12\%.
\subsection{Results}
Fig.~\ref{demo waymo} presents qualitative results of our out-painted driving world together with the corresponding RGB video generation on the Waymo Open Dataset~\cite{Sun_2020_CVPR_WOD}. From top to bottom, we visualize the generated future frames along the predicted trajectory. The first frame corresponds to the ground-truth observation, which provides the initial appearance conditions such as weather and scene style. The subsequent frames illustrate that our method can generate long-horizon driving videos with realistic appearance while maintaining strong controllability.

Fig.~\ref{multi-modal nuscenes} further demonstrates the multi-modal consistency of our framework on the nuScenes dataset~\cite{caesar2020nuscenesmultimodaldatasetautonomous}. Our generated results are aligned across multiple modalities, including RGB images, semantics, depth, occupancy, and LiDAR, highlighting the effectiveness of the proposed 2D–3D coupled generation paradigm.

Fig.~\ref{Fig:sketch} shows that, given an arbitrary user-defined road layout sketch, our framework generates a long-range occupancy scene that follows the specified layout and guides the subsequent video generation.
Tab.~\ref{tab: main} provides a quantitative comparison with existing methods. Despite using only a single multi-view frame as input, InfiniVerse achieves state-of-the-art video generation quality. Thanks to the proposed reciprocal refinement loop, our framework is able to produce the first long-horizon driving videos that remain consistently aligned with the underlying 3D occupancy representation. 

We report a horizon-dependent FID analysis on Waymo. Since Vista~\cite{gao2024vistageneralizabledrivingworld} does not support multi-view generation, we perform this comparison in the single-view setting. For each target horizon $T$, we evaluate the generated frame at step $T{+}1$ and plot FID as a function of rollout length up to 300 frames.

We compare InfiniVerse against Vista in this analysis. Vista is re-implemented by ourselves. As shown in Fig.~\ref{fig:fid_vs_frames}, the FID of InfiniVerse increases much more slowly than that of Vista as the rollout horizon becomes longer.
As shown in Fig.~\ref{fig:fvd_vs_frames}, we report FVD at rollout horizons of 12, 24, 36, 48, 96, and up to 300 frames. As the rollout horizon increases, InfiniVerse consistently maintains lower FVD values and exhibits a slower degradation trend, indicating stronger long-range temporal stability. 
Table~\ref{tab:fvd_vista} presents a direct FVD comparison with Vista~\cite{gao2024vistageneralizabledrivingworld} under different rollout lengths. Among existing baselines, Vista is the only method that can be both reproduced and adapted to a setting close to ours.

We further conduct extensive ablation studies to evaluate the effectiveness of each component in the proposed pipeline in Tab.~\ref{ablation framework}. Using only the 3D branch (voxel-to-video) or only the image branch (image-to-video) results in noticeable performance degradation due to the lack of either temporal consistency or geometric priors. Starting from the reference image and incorporating the 3D occupancy representation, the introduction of the proposed reciprocal loop and region loss significantly improves generation quality, validating the effectiveness of our design.

\section{Conclusion}
In this paper, we introduced \textbf{InfiniVerse}, a unified framework for controllable long-horizon generation of dynamic urban scenes for autonomous driving. Starting from a single multi-view observation, our method reconstructs a 3D occupancy representation that serves as a geometric foundation for autoregressive scene expansion along arbitrary trajectories. 

To bridge spatial reasoning and visual realism, we couple occupancy-guided world modeling with a video diffusion generator. Central to our framework is a \textit{sketch-and-refine} paradigm that establishes a reciprocal feedback loop between 3D occupancy generation and 2D video synthesis: coarse occupancy grids provide structural guidance for video generation, while the generated frames are re-projected to refine the underlying 3D scene representation.

This bidirectional interaction enables consistent geometry–appearance alignment and significantly improves long-horizon stability while reducing hallucination artifacts. Extensive experiments demonstrate that InfiniVerse generates temporally consistent and visually realistic driving scenarios, achieving state-of-the-art performance on standard benchmarks.

Overall, InfiniVerse provides a scalable paradigm for synthesizing controllable urban environments, offering a promising direction for data generation, simulation, and large-scale evaluation in autonomous driving systems.

\clearpage  


%
%
\bibliographystyle{splncs04}
\bibliography{main}

@String(CVPR  = {IEEE Conf. Comput. Vis. Pattern Recog.})

@String(ICLR  = {Int. Conf. Learn. Represent.})

@String(CVPR  = {CVPR})

@String(ICLR  = {ICLR})

@String(CVPR= {IEEE Conf. Comput. Vis. Pattern Recog.})

@String(ICLR = {Int. Conf. Learn. Represent.})

@inproceedings{ren2024xcube,
      title={XCube: Large-Scale 3D Generative Modeling using Sparse Voxel Hierarchies}, 
      author={Ren, Xuanchi and Huang, Jiahui and Zeng, Xiaohui and Museth, Ken 
          and Fidler, Sanja and Williams, Francis},
      booktitle={Proceedings of the IEEE/CVF Conference on Computer Vision and Pattern Recognition},
      year={2024}
      }

@inproceedings{
      ren2024scube,
      title={SCube: Instant Large-Scale Scene Reconstruction using VoxSplats},
      author={Ren, Xuanchi and Lu, Yifan and Liang, Hanxue and Wu, Jay Zhangjie and 
        Ling, Huan and Chen, Mike and Fidler, Sanja annd Williams, Francis and Huang, Jiahui},
      booktitle={The Thirty-eighth Annual Conference on Neural Information Processing Systems},
      year={2024},
    }

@misc{lu2024infinicube,
    title={InfiniCube: Unbounded and Controllable Dynamic 3D Driving Scene Generation with World-Guided Video 
      Models}, 
    author={Yifan Lu and Xuanchi Ren and Jiawei Yang and Tianchang Shen and Zhangjie Wu and Jun Gao and 
      Yue Wang and Siheng Chen and Mike Chen and Sanja Fidler and Jiahui Huang},
    year={2024},
    eprint={2412.03934},
    archivePrefix={arXiv},
    primaryClass={cs.CV},
    url={https://arxiv.org/abs/2412.03934}, 
}

@misc{dosovitskiy2017carlaopenurbandriving,
      title={CARLA: An Open Urban Driving Simulator}, 
      author={Alexey Dosovitskiy and German Ros and Felipe Codevilla and Antonio Lopez and Vladlen Koltun},
      year={2017},
      eprint={1711.03938},
      archivePrefix={arXiv},
      primaryClass={cs.LG},
      url={https://arxiv.org/abs/1711.03938}, 
}

@inproceedings{lee2024semcity,
    title={SemCity: Semantic Scene Generation with Triplane Diffusion},
    author={Lee, Jumin and Lee, Sebin and Jo, Changho and Im, Woobin and Seon, Juhyeong and Yoon, Sung-Eui},
    booktitle={Proceedings of the IEEE/CVF conference on computer vision and pattern recognition},
    year={2024}
}

@misc{liu2024pyramiddiffusionfine3d,
      title={Pyramid Diffusion for Fine 3D Large Scene Generation}, 
      author={Yuheng Liu and Xinke Li and Xueting Li and Lu Qi and Chongshou Li and Ming-Hsuan Yang},
      year={2024},
      eprint={2311.12085},
      archivePrefix={arXiv},
      primaryClass={cs.CV},
      url={https://arxiv.org/abs/2311.12085}, 
}

@article{blockfusion,
  title={BlockFusion: Expandable 3D Scene Generation using Latent Tri-plane Extrapolation},
  author={Wu, Zhennan and Li, Yang and Yan, Han and Shang, Taizhang and Sun, Weixuan and Wang, Senbo and Cui, Ruikai and Liu, Weizhe and Sato, Hiroyuki and Li, Hongdong and Ji, Pan},
  journal={ACM Transactions on Graphics},
  volume={43},
  number={4},
  year={2024},
  doi={10.1145/3658188}
 }

@article{li2024uniscene,
  title={UniScene: Unified Occupancy-centric Driving Scene Generation},
  author={Li, Bohan and Guo, Jiazhe and Liu, Hongsi and Zou, Yingshuang and Ding, Yikang and Chen, Xiwu and Zhu, Hu and Tan, Feiyang and Zhang, Chi and Wang, Tiancai and others},
  journal={arXiv preprint arXiv:2412.05435},
  year={2024}
}

@misc{gao2024vistageneralizabledrivingworld,
      title={Vista: A Generalizable Driving World Model with High Fidelity and Versatile Controllability}, 
      author={Shenyuan Gao and Jiazhi Yang and Li Chen and Kashyap Chitta and Yihang Qiu and Andreas Geiger and Jun Zhang and Hongyang Li},
      year={2024},
      eprint={2405.17398},
      archivePrefix={arXiv},
      primaryClass={cs.CV},
      url={https://arxiv.org/abs/2405.17398}, 
}

@inproceedings{gao2023magicdrive,
  title={{MagicDrive}: Street View Generation with Diverse 3D Geometry Control},
  author={Gao, Ruiyuan and Chen, Kai and Xie, Enze and Hong, Lanqing and Li, Zhenguo and Yeung, Dit-Yan and Xu, Qiang},
  booktitle = {International Conference on Learning Representations},
  year={2024}
}

@misc{gao2025magicdrive3dcontrollable3dgeneration,
      title={MagicDrive3D: Controllable 3D Generation for Any-View Rendering in Street Scenes}, 
      author={Ruiyuan Gao and Kai Chen and Zhihao Li and Lanqing Hong and Zhenguo Li and Qiang Xu},
      year={2025},
      eprint={2405.14475},
      archivePrefix={arXiv},
      primaryClass={cs.CV},
      url={https://arxiv.org/abs/2405.14475}, 
}

@misc{lin2023infinicityinfinitescalecitysynthesis,
      title={InfiniCity: Infinite-Scale City Synthesis}, 
      author={Chieh Hubert Lin and Hsin-Ying Lee and Willi Menapace and Menglei Chai and Aliaksandr Siarohin and Ming-Hsuan Yang and Sergey Tulyakov},
      year={2023},
      eprint={2301.09637},
      archivePrefix={arXiv},
      primaryClass={cs.CV},
      url={https://arxiv.org/abs/2301.09637}, 
}

@misc{lu2024wovogenworldvolumeawarediffusion,
      title={WoVoGen: World Volume-aware Diffusion for Controllable Multi-camera Driving Scene Generation}, 
      author={Jiachen Lu and Ze Huang and Zeyu Yang and Jiahui Zhang and Li Zhang},
      year={2024},
      eprint={2312.02934},
      archivePrefix={arXiv},
      primaryClass={cs.CV},
      url={https://arxiv.org/abs/2312.02934}, 
}

@misc{yang2025cogvideoxtexttovideodiffusionmodels,
      title={CogVideoX: Text-to-Video Diffusion Models with An Expert Transformer}, 
      author={Zhuoyi Yang and Jiayan Teng and Wendi Zheng and Ming Ding and Shiyu Huang and Jiazheng Xu and Yuanming Yang and Wenyi Hong and Xiaohan Zhang and Guanyu Feng and Da Yin and Yuxuan Zhang and Weihan Wang and Yean Cheng and Bin Xu and Xiaotao Gu and Yuxiao Dong and Jie Tang},
      year={2025},
      eprint={2408.06072},
      archivePrefix={arXiv},
      primaryClass={cs.CV},
      url={https://arxiv.org/abs/2408.06072}, 
}

@misc{blattmann2023stablevideodiffusionscaling,
      title={Stable Video Diffusion: Scaling Latent Video Diffusion Models to Large Datasets}, 
      author={Andreas Blattmann and Tim Dockhorn and Sumith Kulal and Daniel Mendelevitch and Maciej Kilian and Dominik Lorenz and Yam Levi and Zion English and Vikram Voleti and Adam Letts and Varun Jampani and Robin Rombach},
      year={2023},
      eprint={2311.15127},
      archivePrefix={arXiv},
      primaryClass={cs.CV},
      url={https://arxiv.org/abs/2311.15127}, 
}

@article{ye2025hi3dgen,
  title={Hi3DGen: High-fidelity 3D Geometry Generation from Images via Normal Bridging},
  author={Ye, Chongjie and Wu, Yushuang and Lu, Ziteng and Chang, Jiahao and Guo, Xiaoyang and Zhou, Jiaqing and Zhao, Hao and Han, Xiaoguang},
  journal={arXiv preprint arXiv:2503.22236}, 
  year={2025}
}

@misc{graham2017submanifoldsparseconvolutionalnetworks,
      title={Submanifold Sparse Convolutional Networks}, 
      author={Benjamin Graham and Laurens van der Maaten},
      year={2017},
      eprint={1706.01307},
      archivePrefix={arXiv},
      primaryClass={cs.NE},
      url={https://arxiv.org/abs/1706.01307}, 
}

@misc{philion2020liftsplatshootencoding,
      title={Lift, Splat, Shoot: Encoding Images From Arbitrary Camera Rigs by Implicitly Unprojecting to 3D}, 
      author={Jonah Philion and Sanja Fidler},
      year={2020},
      eprint={2008.05711},
      archivePrefix={arXiv},
      primaryClass={cs.CV},
      url={https://arxiv.org/abs/2008.05711}, 
}

@misc{oquab2024dinov2learningrobustvisual,
      title={DINOv2: Learning Robust Visual Features without Supervision}, 
      author={Maxime Oquab and Timothée Darcet and Théo Moutakanni and Huy Vo and Marc Szafraniec and Vasil Khalidov and Pierre Fernandez and Daniel Haziza and Francisco Massa and Alaaeldin El-Nouby and Mahmoud Assran and Nicolas Ballas and Wojciech Galuba and Russell Howes and Po-Yao Huang and Shang-Wen Li and Ishan Misra and Michael Rabbat and Vasu Sharma and Gabriel Synnaeve and Hu Xu and Hervé Jegou and Julien Mairal and Patrick Labatut and Armand Joulin and Piotr Bojanowski},
      year={2024},
      eprint={2304.07193},
      archivePrefix={arXiv},
      primaryClass={cs.CV},
      url={https://arxiv.org/abs/2304.07193}, 
}

@misc{liu2023one2345fastsingleimage,
      title={One-2-3-45++: Fast Single Image to 3D Objects with Consistent Multi-View Generation and 3D Diffusion}, 
      author={Minghua Liu and Ruoxi Shi and Linghao Chen and Zhuoyang Zhang and Chao Xu and Xinyue Wei and Hansheng Chen and Chong Zeng and Jiayuan Gu and Hao Su},
      year={2023},
      eprint={2311.07885},
      archivePrefix={arXiv},
      primaryClass={cs.CV},
      url={https://arxiv.org/abs/2311.07885}, 
}

@misc{sitzmann2019deepvoxelslearningpersistent3d,
      title={DeepVoxels: Learning Persistent 3D Feature Embeddings}, 
      author={Vincent Sitzmann and Justus Thies and Felix Heide and Matthias Nießner and Gordon Wetzstein and Michael Zollhöfer},
      year={2019},
      eprint={1812.01024},
      archivePrefix={arXiv},
      primaryClass={cs.CV},
      url={https://arxiv.org/abs/1812.01024}, 
}

@misc{sun2021neuralreconrealtimecoherent3d,
      title={NeuralRecon: Real-Time Coherent 3D Reconstruction from Monocular Video}, 
      author={Jiaming Sun and Yiming Xie and Linghao Chen and Xiaowei Zhou and Hujun Bao},
      year={2021},
      eprint={2104.00681},
      archivePrefix={arXiv},
      primaryClass={cs.CV},
      url={https://arxiv.org/abs/2104.00681}, 
}

@misc{gao2024cat3dcreate3dmultiview,
      title={CAT3D: Create Anything in 3D with Multi-View Diffusion Models}, 
      author={Ruiqi Gao and Aleksander Holynski and Philipp Henzler and Arthur Brussee and Ricardo Martin-Brualla and Pratul Srinivasan and Jonathan T. Barron and Ben Poole},
      year={2024},
      eprint={2405.10314},
      archivePrefix={arXiv},
      primaryClass={cs.CV},
      url={https://arxiv.org/abs/2405.10314}, 
}

@misc{kim2023neuralfieldldmscenegenerationhierarchical,
      title={NeuralField-LDM: Scene Generation with Hierarchical Latent Diffusion Models}, 
      author={Seung Wook Kim and Bradley Brown and Kangxue Yin and Karsten Kreis and Katja Schwarz and Daiqing Li and Robin Rombach and Antonio Torralba and Sanja Fidler},
      year={2023},
      eprint={2304.09787},
      archivePrefix={arXiv},
      primaryClass={cs.CV},
      url={https://arxiv.org/abs/2304.09787}, 
}

@misc{zhang2024urbanscenediffusionsemantic,
      title={Urban Scene Diffusion through Semantic Occupancy Map}, 
      author={Junge Zhang and Qihang Zhang and Li Zhang and Ramana Rao Kompella and Gaowen Liu and Bolei Zhou},
      year={2024},
      eprint={2403.11697},
      archivePrefix={arXiv},
      primaryClass={cs.CV},
      url={https://arxiv.org/abs/2403.11697}, 
}

@misc{zyrianov2024lidardmgenerativelidarsimulation,
      title={LidarDM: Generative LiDAR Simulation in a Generated World}, 
      author={Vlas Zyrianov and Henry Che and Zhijian Liu and Shenlong Wang},
      year={2024},
      eprint={2404.02903},
      archivePrefix={arXiv},
      primaryClass={cs.CV},
      url={https://arxiv.org/abs/2404.02903}, 
}

@misc{shi2024mvdreammultiviewdiffusion3d,
      title={MVDream: Multi-view Diffusion for 3D Generation}, 
      author={Yichun Shi and Peng Wang and Jianglong Ye and Mai Long and Kejie Li and Xiao Yang},
      year={2024},
      eprint={2308.16512},
      archivePrefix={arXiv},
      primaryClass={cs.CV},
      url={https://arxiv.org/abs/2308.16512}, 
}

@misc{xu2024grmlargegaussianreconstruction,
      title={GRM: Large Gaussian Reconstruction Model for Efficient 3D Reconstruction and Generation}, 
      author={Yinghao Xu and Zifan Shi and Wang Yifan and Hansheng Chen and Ceyuan Yang and Sida Peng and Yujun Shen and Gordon Wetzstein},
      year={2024},
      eprint={2403.14621},
      archivePrefix={arXiv},
      primaryClass={cs.CV},
      url={https://arxiv.org/abs/2403.14621}, 
}

@InProceedings{Sun_2020_CVPR_WOD, author = {Sun, Pei and Kretzschmar, Henrik and Dotiwalla, Xerxes and Chouard, Aurelien and Patnaik, Vijaysai and Tsui, Paul and Guo, James and Zhou, Yin and Chai, Yuning and Caine, Benjamin and Vasudevan, Vijay and Han, Wei and Ngiam, Jiquan and Zhao, Hang and Timofeev, Aleksei and Ettinger, Scott and Krivokon, Maxim and Gao, Amy and Joshi, Aditya and Zhang, Yu and Shlens, Jonathon and Chen, Zhifeng and Anguelov, Dragomir}, title = {Scalability in Perception for Autonomous Driving: Waymo Open Dataset}, booktitle = {Proceedings of the IEEE/CVF Conference on Computer Vision and Pattern Recognition (CVPR)}, month = {June}, year = {2020} }

@misc{caesar2020nuscenesmultimodaldatasetautonomous,
      title={nuScenes: A multimodal dataset for autonomous driving}, 
      author={Holger Caesar and Varun Bankiti and Alex H. Lang and Sourabh Vora and Venice Erin Liong and Qiang Xu and Anush Krishnan and Yu Pan and Giancarlo Baldan and Oscar Beijbom},
      year={2020},
      eprint={1903.11027},
      archivePrefix={arXiv},
      primaryClass={cs.LG},
      url={https://arxiv.org/abs/1903.11027}, 
}

@article{chen2024far,
    title={How Far Are We to GPT-4V? Closing the Gap to Commercial Multimodal Models with Open-Source Suites},
    author={Chen, Zhe and Wang, Weiyun and Tian, Hao and Ye, Shenglong and Gao, Zhangwei and Cui, Erfei and Tong, Wenwen and Hu, Kongzhi and Luo, Jiapeng and Ma, Zheng and others},
    journal={arXiv preprint arXiv:2404.16821},
    year={2024}
  }

@inproceedings{chen2024internvl,
    title={Internvl: Scaling up vision foundation models and aligning for generic visual-linguistic tasks},
    author={Chen, Zhe and Wu, Jiannan and Wang, Wenhai and Su, Weijie and Chen, Guo and Xing, Sen and Zhong, Muyan and Zhang, Qinglong and Zhu, Xizhou and Lu, Lewei and others},
    booktitle={Proceedings of the IEEE/CVF Conference on Computer Vision and Pattern Recognition},
    pages={24185--24198},
    year={2024}
  }

@misc{wang2024videoclipxladvancinglongdescription,
      title={VideoCLIP-XL: Advancing Long Description Understanding for Video CLIP Models}, 
      author={Jiapeng Wang and Chengyu Wang and Kunzhe Huang and Jun Huang and Lianwen Jin},
      year={2024},
      eprint={2410.00741},
      archivePrefix={arXiv},
      primaryClass={cs.CL},
      url={https://arxiv.org/abs/2410.00741}, 
}

@misc{heusel2018ganstrainedtimescaleupdate,
      title={GANs Trained by a Two Time-Scale Update Rule Converge to a Local Nash Equilibrium}, 
      author={Martin Heusel and Hubert Ramsauer and Thomas Unterthiner and Bernhard Nessler and Sepp Hochreiter},
      year={2018},
      eprint={1706.08500},
      archivePrefix={arXiv},
      primaryClass={cs.LG},
      url={https://arxiv.org/abs/1706.08500}, 
}

@misc{wang2023videocomposercompositionalvideosynthesis,
      title={VideoComposer: Compositional Video Synthesis with Motion Controllability}, 
      author={Xiang Wang and Hangjie Yuan and Shiwei Zhang and Dayou Chen and Jiuniu Wang and Yingya Zhang and Yujun Shen and Deli Zhao and Jingren Zhou},
      year={2023},
      eprint={2306.02018},
      archivePrefix={arXiv},
      primaryClass={cs.CV},
      url={https://arxiv.org/abs/2306.02018}, 
}

@article{sima2023_occnet,
    title={Scene as Occupancy},
    author={Chonghao Sima and Wenwen Tong and Tai Wang and Li Chen and Silei Wu and Hanming Deng and Yi Gu and Lewei Lu and Ping Luo and Dahua Lin and Hongyang Li},
    year={2023},
    eprint={2306.02851},
    archivePrefix={arXiv},
    primaryClass={cs.CV}
}

@misc{chen2023voxelnextfullysparsevoxelnet,
      title={VoxelNeXt: Fully Sparse VoxelNet for 3D Object Detection and Tracking}, 
      author={Yukang Chen and Jianhui Liu and Xiangyu Zhang and Xiaojuan Qi and Jiaya Jia},
      year={2023},
      eprint={2303.11301},
      archivePrefix={arXiv},
      primaryClass={cs.CV},
      url={https://arxiv.org/abs/2303.11301}, 
}

@misc{xie2024citydreamercompositionalgenerativemodel,
      title={CityDreamer: Compositional Generative Model of Unbounded 3D Cities}, 
      author={Haozhe Xie and Zhaoxi Chen and Fangzhou Hong and Ziwei Liu},
      year={2024},
      eprint={2309.00610},
      archivePrefix={arXiv},
      primaryClass={cs.CV},
      url={https://arxiv.org/abs/2309.00610}, 
}

@misc{tian2023occ3dlargescale3doccupancy,
      title={Occ3D: A Large-Scale 3D Occupancy Prediction Benchmark for Autonomous Driving}, 
      author={Xiaoyu Tian and Tao Jiang and Longfei Yun and Yucheng Mao and Huitong Yang and Yue Wang and Yilun Wang and Hang Zhao},
      year={2023},
      eprint={2304.14365},
      archivePrefix={arXiv},
      primaryClass={cs.CV},
      url={https://arxiv.org/abs/2304.14365}, 
}

@article{llama3modelcard,

title={Llama 3 Model Card},

author={AI@Meta},

year={2024},

url = {https://github.com/meta-llama/llama3/blob/main/MODEL_CARD.md}

}

@misc{zhao2023unipcunifiedpredictorcorrectorframework,
      title={UniPC: A Unified Predictor-Corrector Framework for Fast Sampling of Diffusion Models}, 
      author={Wenliang Zhao and Lujia Bai and Yongming Rao and Jie Zhou and Jiwen Lu},
      year={2023},
      eprint={2302.04867},
      archivePrefix={arXiv},
      primaryClass={cs.LG},
      url={https://arxiv.org/abs/2302.04867}, 
}

@misc{ho2022classifierfreediffusionguidance,
      title={Classifier-Free Diffusion Guidance}, 
      author={Jonathan Ho and Tim Salimans},
      year={2022},
      eprint={2207.12598},
      archivePrefix={arXiv},
      primaryClass={cs.LG},
      url={https://arxiv.org/abs/2207.12598}, 
}

@inproceedings{schoenberger2016sfm,
    author={Sch\"{o}nberger, Johannes Lutz and Frahm, Jan-Michael},
    title={Structure-from-Motion Revisited},
    booktitle={Conference on Computer Vision and Pattern Recognition (CVPR)},
    year={2016},
}

@misc{kim2021drivegancontrollablehighqualityneural,
      title={DriveGAN: Towards a Controllable High-Quality Neural Simulation}, 
      author={Seung Wook Kim and Jonah Philion and Antonio Torralba and Sanja Fidler},
      year={2021},
      eprint={2104.15060},
      archivePrefix={arXiv},
      primaryClass={cs.CV},
      url={https://arxiv.org/abs/2104.15060}, 
}

@misc{yang2025xscenelargescaledrivingscene,
      title={X-Scene: Large-Scale Driving Scene Generation with High Fidelity and Flexible Controllability}, 
      author={Yu Yang and Alan Liang and Jianbiao Mei and Yukai Ma and Yong Liu and Gim Hee Lee},
      year={2025},
      eprint={2506.13558},
      archivePrefix={arXiv},
      primaryClass={cs.CV},
      url={https://arxiv.org/abs/2506.13558}, 
}

@misc{zhou2018stereomagnificationlearningview,
      title={Stereo Magnification: Learning View Synthesis using Multiplane Images}, 
      author={Tinghui Zhou and Richard Tucker and John Flynn and Graham Fyffe and Noah Snavely},
      year={2018},
      eprint={1805.09817},
      archivePrefix={arXiv},
      primaryClass={cs.CV},
      url={https://arxiv.org/abs/1805.09817}, 
}

@misc{gao2025magicdrivev2highresolutionlongvideo,
      title={MagicDrive-V2: High-Resolution Long Video Generation for Autonomous Driving with Adaptive Control}, 
      author={Ruiyuan Gao and Kai Chen and Bo Xiao and Lanqing Hong and Zhenguo Li and Qiang Xu},
      year={2025},
      eprint={2411.13807},
      archivePrefix={arXiv},
      primaryClass={cs.CV},
      url={https://arxiv.org/abs/2411.13807}, 
}

@misc{swerdlow2024streetviewimagegenerationbirdseye,
      title={Street-View Image Generation from a Bird's-Eye View Layout}, 
      author={Alexander Swerdlow and Runsheng Xu and Bolei Zhou},
      year={2024},
      eprint={2301.04634},
      archivePrefix={arXiv},
      primaryClass={cs.CV},
      url={https://arxiv.org/abs/2301.04634}, 
}

@misc{yang2023bevcontrolaccuratelycontrollingstreetview,
      title={BEVControl: Accurately Controlling Street-view Elements with Multi-perspective Consistency via BEV Sketch Layout}, 
      author={Kairui Yang and Enhui Ma and Jibin Peng and Qing Guo and Di Lin and Kaicheng Yu},
      year={2023},
      eprint={2308.01661},
      archivePrefix={arXiv},
      primaryClass={cs.CV},
      url={https://arxiv.org/abs/2308.01661}, 
}

@misc{wang2023drivedreamerrealworlddrivenworldmodels,
      title={DriveDreamer: Towards Real-world-driven World Models for Autonomous Driving}, 
      author={Xiaofeng Wang and Zheng Zhu and Guan Huang and Xinze Chen and Jiagang Zhu and Jiwen Lu},
      year={2023},
      eprint={2309.09777},
      archivePrefix={arXiv},
      primaryClass={cs.CV},
      url={https://arxiv.org/abs/2309.09777}, 
}

@misc{wen2023panaceapanoramiccontrollablevideo,
      title={Panacea: Panoramic and Controllable Video Generation for Autonomous Driving}, 
      author={Yuqing Wen and Yucheng Zhao and Yingfei Liu and Fan Jia and Yanhui Wang and Chong Luo and Chi Zhang and Tiancai Wang and Xiaoyan Sun and Xiangyu Zhang},
      year={2023},
      eprint={2311.16813},
      archivePrefix={arXiv},
      primaryClass={cs.CV},
      url={https://arxiv.org/abs/2311.16813}, 
}

@misc{zhao2024drivedreamer2llmenhancedworldmodels,
      title={DriveDreamer-2: LLM-Enhanced World Models for Diverse Driving Video Generation}, 
      author={Guosheng Zhao and Xiaofeng Wang and Zheng Zhu and Xinze Chen and Guan Huang and Xiaoyi Bao and Xingang Wang},
      year={2024},
      eprint={2403.06845},
      archivePrefix={arXiv},
      primaryClass={cs.CV},
      url={https://arxiv.org/abs/2403.06845}, 
}

@misc{yang2024genadgeneralizedpredictivemodel,
      title={GenAD: Generalized Predictive Model for Autonomous Driving}, 
      author={Jiazhi Yang and Shenyuan Gao and Yihang Qiu and Li Chen and Tianyu Li and Bo Dai and Kashyap Chitta and Penghao Wu and Jia Zeng and Ping Luo and Jun Zhang and Andreas Geiger and Yu Qiao and Hongyang Li},
      year={2024},
      eprint={2403.09630},
      archivePrefix={arXiv},
      primaryClass={cs.CV},
      url={https://arxiv.org/abs/2403.09630}, 
}

@misc{jiang2025diveefficientmultiviewdriving,
      title={DiVE: Efficient Multi-View Driving Scenes Generation Based on Video Diffusion Transformer}, 
      author={Junpeng Jiang and Gangyi Hong and Miao Zhang and Hengtong Hu and Kun Zhan and Rui Shao and Liqiang Nie},
      year={2025},
      eprint={2504.19614},
      archivePrefix={arXiv},
      primaryClass={cs.CV},
      url={https://arxiv.org/abs/2504.19614}, 
}

@misc{wang2023drivingfuturemultiviewvisual,
      title={Driving into the Future: Multiview Visual Forecasting and Planning with World Model for Autonomous Driving}, 
      author={Yuqi Wang and Jiawei He and Lue Fan and Hongxin Li and Yuntao Chen and Zhaoxiang Zhang},
      year={2023},
      eprint={2311.17918},
      archivePrefix={arXiv},
      primaryClass={cs.CV},
      url={https://arxiv.org/abs/2311.17918}, 
}

@misc{ma2024unleashinggeneralizationendtoendautonomous,
      title={Unleashing Generalization of End-to-End Autonomous Driving with Controllable Long Video Generation}, 
      author={Enhui Ma and Lijun Zhou and Tao Tang and Zhan Zhang and Dong Han and Junpeng Jiang and Kun Zhan and Peng Jia and Xianpeng Lang and Haiyang Sun and Di Lin and Kaicheng Yu},
      year={2024},
      eprint={2406.01349},
      archivePrefix={arXiv},
      primaryClass={cs.CV},
      url={https://arxiv.org/abs/2406.01349}, 
}

@inproceedings{DBLP:conf/iclr/UnterthinerSKMM19,
  author       = {Thomas Unterthiner and
                  Sjoerd van Steenkiste and
                  Karol Kurach and
                  Rapha{\"{e}}l Marinier and
                  Marcin Michalski and
                  Sylvain Gelly},
  title        = {{FVD:} {A} new Metric for Video Generation},
  booktitle    = {Deep Generative Models for Highly Structured Data, {ICLR} 2019 Workshop,
                  New Orleans, Louisiana, United States, May 6, 2019},
  publisher    = {OpenReview.net},
  year         = {2019},
  url          = {https://openreview.net/forum?id=rylgEULtdN},
  bibsource    = {dblp computer science bibliography, https://dblp.org}
}

@misc{rombach2022highresolutionimagesynthesislatent,
      title={High-Resolution Image Synthesis with Latent Diffusion Models}, 
      author={Robin Rombach and Andreas Blattmann and Dominik Lorenz and Patrick Esser and Björn Ommer},
      year={2022},
      eprint={2112.10752},
      archivePrefix={arXiv},
      primaryClass={cs.CV},
      url={https://arxiv.org/abs/2112.10752}, 
}
\end{document}